\documentclass{article}

\usepackage[final,nonatbib]{icbinb2021}

\usepackage[numbers]{natbib}

\usepackage[utf8]{inputenc} 
\usepackage[T1]{fontenc}    
\usepackage{url}            
\usepackage{booktabs}       
\usepackage{amsfonts}       
\usepackage{nicefrac}       
\usepackage{microtype}     
\usepackage{subcaption}

\usepackage{float}
\usepackage{graphicx}
\usepackage{caption}
\usepackage{subcaption}
\usepackage{hyperref}

\usepackage{longtable}
\usepackage{amsfonts}
\usepackage{booktabs}
\usepackage[dvipsnames]{xcolor}

\usepackage{amsmath}
\usepackage{amssymb}
\usepackage{multirow}
\usepackage{dsfont}
\usepackage{xcolor,colortbl}
\usepackage{mathtools}
\usepackage{wrapfig}

\usepackage[utf8]{inputenc}    
\usepackage{url}            
\usepackage{booktabs}  
\usepackage{nicefrac}       
\usepackage{microtype} 

\usepackage{algorithm}
\usepackage{listings}

\usepackage[english]{babel}
\usepackage{amsthm}

\newcommand{\algcomment}[1]{%
    \vspace{4pt}%
    \noindent%
    {\footnotesize #1\par}%
    \vspace{\baselineskip}%
    }

\newtheorem{innercustomthm}{Theorem}
\newenvironment{customthm}[1]
  {\renewcommand\theinnercustomthm{#1}\innercustomthm}
  {\endinnercustomthm}

\title{Improvising the Learning of Neural Networks on Hyperspherical Manifold}

\author{%
  Lalith Bharadwaj Baru\thanks{The authors contributed equally}\\
  VNRVJIET\\
  Hyderabad-90, T.S, India.\\
  \texttt{lalithbharadwaj313@gmail.com}\\ 
  \And
  Sai Vardhan Kanumolu$^*$\\
  VNRVJIET\\
  Hyderabad-90, T.S, India.\\
  \texttt{kanumolusaivardhan@gmail.com} \\
  \And
  Akshay Patel Shilhora$^*$\\
  VNRVJIET\\
  Hyderabad-90, T.S, India.\\
  \texttt{shilhora.akshay333@gmail.com} \\
  \AND
  Madhu G\\
  VNRVJIET\\
  Hyderabad-90, T.S, India.\\
  \texttt{madhu\_g@vnrvjiet.in} \\
}

\begin{document}

\maketitle

\begin{abstract}
The impact of convolution neural networks (CNNs) in the supervised settings provided tremendous increment in performance. The representations learned from CNN's operated on hyperspherical manifold led to insightful outcomes in face recognition, face identification, and other supervised tasks. A broad range of activation functions were developed with hypersphere intuition which performs superior to softmax in euclidean space. The main motive of this research is to provide insights. First, the stereographic projection is implied to transform data from Euclidean space ($\mathbb{R}^{n}$) to hyperspherical manifold ($\mathbb{S}^{n}$) to analyze the performance of angular margin losses. Secondly, proving theoretically and practically that decision boundaries constructed on hypersphere using stereographic projection obliges the learning of neural networks. Experiments have demonstrated that applying stereographic projection on existing state-of-the-art angular margin objective functions improved performance for standard image classification data sets (CIFAR-10,100). Further, we ran our experiments on malaria-thin blood smear images, resulting in effective outcomes. The code is publicly available at:
: \textbf{\url{https://github.com/barulalithb/stereo-angular-margin}}.
\end{abstract}

\section{Introduction}
The impact of deep learning has provided a notable surge in various domains such as computer vision, language processing, speech processing, and graph mining \citep{lecun2015deep}. A wide range of neural networks is developed by captivating their performance in deep learning. Specifically, convolution neural networks (CNNs) have shown their impressive ability to extract invariances in many problems. In computer vision, the CNN's were implied to extract features that are class specific, hierarchical, and other complex invariances ~\citep{zeiler2014visualizing}. Hence, these CNNs are used as \textit{encoders} which map high dimensional representations to lower dimensions (feature sets). 


Neural networks which were devised on hyperspherical manifold provided significant performance compared to standard CNNs ~\citep{deephyper1}~\citep{deephyper2}. A broad range of objective functions was devised for solving face verification, face recognition, and similar supervised tasks with hyperspherical intuition. Also, this intuition is carried by the recent advances in self-supervision by learning features contrastively. Further, stereographic projection (SP) on hyperspheres is utilized as a pre-processing technique to understand continuous rotational invariances. 

It is observed that erstwhile research (which specifically operates on hyperspherical manifold) assumes that their objective functions are operated on the hypersphere. But in this work, we utilize a geometric transformation to map the data points from the euclidean to hyperspherical decision region. Further, we understand the behavior of angular margin function and standard categorical cross-entropy (CCE) operated on this manifold.

\subsection{Impact of angular margin losses}
In recent years various losses have been stated for specific objectives. Rather than learning separable
features, the discriminative approach encourages us to learn features selectively, thus increasing
compactness in intra-class and separability in inter-class features. In discriminative learning, research
has shown that cosine-margin for learned features has significant improvement than euclidean-
margin. Many angular-margin losses have been implemented, specifically for the face recognition task, based on the analogy that the learned
features from a neural network interpret the geometry as a hyperspherical manifold. Where a margin can control the decision
boundary. ~\citep{liu2016large} has developed the motive of introducing angular
constraints to decision functions. Further,~\citep{liu2017sphereface} has shown normalizing the weight vectors will advance
the  ~\citep{liu2016large} with no requirement of joint-supervision by a straightforward implementation ~\citep{wang2018cosface} ~\citep{deng2019arcface} has improved
the performance of classifiers by introducing additive-margin loss function, thus tuning the cosine
space and angle space with the margin by rescaling their logits with a fixed norm.~\citep{kim2020broadface} has
overcome the problem of mini-batch learning by adding queues to their framework containing embedding
vectors learned from the neural network with their corresponding identity-representative vector and these
vectors are updated throughout training. Although ~\citep{kim2020broadface} has utilized the loss function presented in ~\citep{deng2019arcface},
it adopted a new training procedure which has proven significant improvement in the face recognition
dataset.

\subsection{Impact on contrastive representation learning}
Recently, contrastive representation learning has advanced in deep understanding to learn invariances both in supervised ~\citep{NEURIPS2020_d89a66c7} and self-supervised setting ~\citep{chen2020simple}. ~\citep{wang2020understanding} produced two crucial factors in designing a contrastive objective function: a) uniformity and b) alignment. These factors played a vital role in developing a resilient contrastive objective function. While learning representations contrastively, ~\citep{NEURIPS2020_d89a66c7} developed a contrastive loss which was superior to that of self-supervised method ~\citep{chen2020simple}. Further, ~\citep{NEURIPS2020_d89a66c7} performed unit sphere normalization to feature sets drawn from an encoder and resulted in outstanding performance. Both the works focused on training dynamics on unit hypersphere by designing a contrastive loss. 

\subsection{Stereographic projections}
Closely related work, implementing stereographic projection,  was done by ~\citep{saffery1991using}, ~\citep{park2019sphere},and ~\citep{zhou2019continuity}. First,~\citep{saffery1991using} utilized stereographic projection as a pre-processing technique to classify 4-bit parity. The projection transforms real-valued patterns from one space to another by increasing the dimensionality and providing superior performance on the two-spiral problem. Where ~\citep{park2019sphere}, proposed SphereGAN for image generation problem by utilizing stereographic projection onto hypersphere to attain state-of-the-art results. ~\citep{zhou2019continuity} tackled with continuity of rotation representations for better training dynamics of neural networks. The insights of stereographic projections are implied while understanding the n-dimensional groups with rotations. ~\citep{yavartanoo2018spnet} have utilized stereographic projection as a pre-processing technique to transform 3D objects to 2D planar images.

The existing research, as mentioned, primarily focused on developing objective functions which assume that data points lie in the hypersphere. Hence, an appropriate map is not provided to shift the space and learn invariances from neural networks. Accordingly, we solve this problem by providing a detailed solution.

\section{Uniqueness of Proposed Work}
The uniqueness of our research is twofold. 
\begin{enumerate}
\item First, we imply stereographic projection to transform feature vectors from euclidean space ($R^n$) to hyperspherical manifold ($S^n$) to analyze the performance of angular margin losses.
\item Second, proving theoretically and practically that decision boundaries constructed on hypersphere using stereographic projection oblige the learning of neural networks.
\end{enumerate}


\section{Stereographic projection on hyperspherical manifold}
\paragraph{Definition} \textit{Stereographic projection is said to be a map $ \varphi: \mathbb{R}^n = \{x \in \mathbb{R}^{n+1} : x_{n+1} =0\} \to \mathbb{S}^n - \{e_{n+1}\}$ of $x$ $((n+1)^{th}$dimensional tuple$)$ in euclidean space onto a hypersphere $(\mathbb{S}^n)$ such that, the x meets a point on hypersphere by linearly extending it till $e_{n+1}$.}

Let $e_i$ be a vector in $\mathbb{R}^n$ where, $i^{th}$ coordinate is unitary and else are null. Then $\{e_1, e_2,..., e_n\}$ is orthonormal basis of $\mathbb{R}^n$ and also called as \textit{standard basis} in $\mathbb{R}^n$. So, $e_{n+1}$ is $(n+1)^{th}$ standard basis in $\mathbb{R}^{n+1}$. So, $\varphi(x)$ is the stereographic projection of $x$, in euclidean manifold, onto the hyperspherical manifold. This $\varphi(x)$ lie on the line passing through x and is directed towards $e_{n+1} - x$. Hence, $ \varphi(x)$ can be written as,
\begin{equation}\label{eq1}
\varphi(x) := x + z*(e_{n+1} - x)
\end{equation}

\begin{figure*}[!b]
\begin{center}
\includegraphics[width=0.75\textwidth]{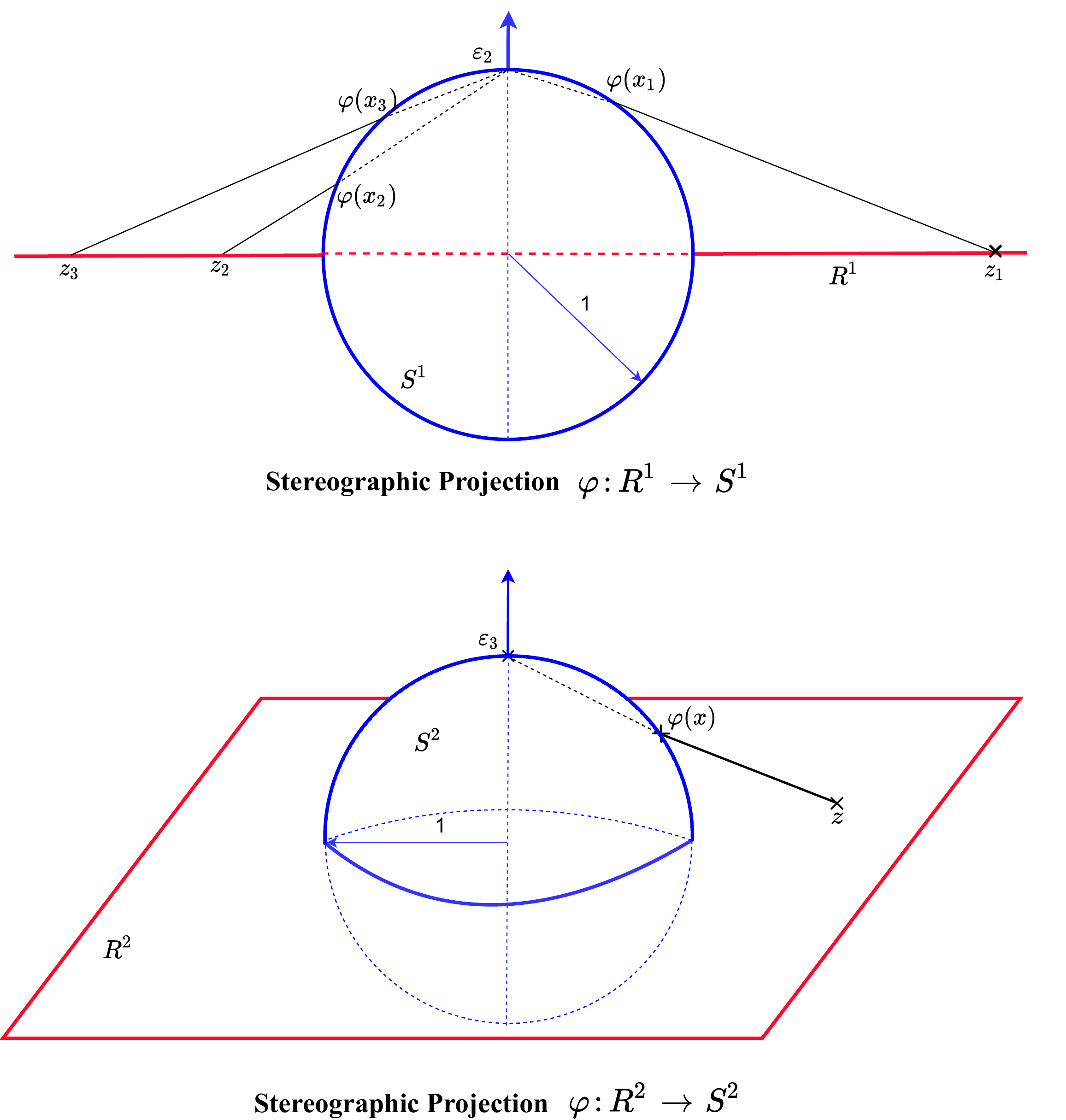}
\caption{Stereographic projections of a point on the real line and a plane are projected onto a circle and a sphere. }\label{fig:sp}
\end{center}
\end{figure*}

A visual illustration of stereographic projection from $\mathbb{R}^{1} \to \mathbb{S}^{1}$ and $\mathbb{R}^{2} \to \mathbb{S}^{2}$  is provided for better intuitiveness in the Fig.~\ref{fig:sp}. Further, there are two underlying key points to be noted while applying stereographic projection are,
\begin{itemize}
\item The euclidean space ($\mathbb{R}^{n}$) must contain the origin.
\item The radius of the hypersphere is unity.
\end{itemize}

As, $ \varphi(x)$ passes through the unit sphere,
\begin{equation}
{\mid\varphi(x)\mid}^2 := 1
\end{equation}

Substituting equation (1) in (2) results,
\begin{equation}
z := \frac{{\mid x \mid}^2 - 1}{{\mid x \mid}^2 + 1}
\end{equation}

Hence, the coordinates obtained after the stereographic projection $ \varphi$ of $x$ ($x:(x_1,x_2,...,x_{n+1})$) onto hypersphere are,

\begin{equation}
\varphi(x) := \bigg( \frac{2x_1}{{\mid x \mid}^2 + 1}, \frac{2x_2}{{\mid x \mid}^2 + 1},...,\frac{{\mid x \mid}^2 - 1}{{\mid x \mid}^2 + 1} \bigg)
\end{equation}



\begin{algorithm*}[t]\label{alg:sp}
\caption{ PyTorch-like pseudo code for stereographic projection}

\label{alg:code}
\algcomment{\fontsize{7.2pt}{0em}\selectfont \texttt{i/p}: input; \texttt{o/p}: output; \texttt{dim}: dimensional ; \texttt{concat}: concatenation; \texttt{sum}: summation;  \texttt{pow(.,2)}: Squaring the input
}

\definecolor{codeblue}{rgb}{0,0.10,0.65}
\definecolor{codered}{rgb}{1,0.2,0}
\definecolor{codegreen}{rgb}{0.1,0.8,0}
\definecolor{backcolour}{rgb}{0.95,0.95,0.92}

\lstset{
  backgroundcolor=\color{white},
  basicstyle=\fontsize{9pt}{9pt}\ttfamily\selectfont,
  columns=fullflexible,
  breaklines=true,
  stringstyle=\color{codegreen},
  captionpos=b,
  commentstyle=\fontsize{7.2pt}{7.2pt}\color{codeblue},
  keywordstyle=\fontsize{7.2pt}{7.2pt}\color{codered},
}

\begin{lstlisting}[language=python]
# This function considers  (n)-dim i/p vector and returns (n+1)-dim o/p vector.
# in_vec: input vector
# norm: norm of in_vec vector
# a: first n dimensions of the projection 
# b: final dimension of the projection
# out: stereographic projection

def streographic_preojection (in_vec):  
    norm = sum(pow(in_vec,2))
    a = (2*in_vec)/(norm+1)
    b = (norm-1)/(norm+1)
    out = concat(a,b)
    return out
    
\end{lstlisting}
\end{algorithm*}

\section{Hypersphere as decision region}
This section theoretically proves that the hypersphere can be used as a decision region in feed-forward neural networks. To obtain the transformation from euclidean space ($\mathbb{R}^{n}$) to hyperspherical manifold ($\mathbb{S}^{n}$), we imply the earlier described concept stereographic projection.
\begin{customthm}{1}\label{theorem1}
A unit hyperspherical manifold $\mathbb{S}^{n}$ is always convex and connected.
\end{customthm}

\begin{proof}
First the convexity and next the connectedness are proven.
\paragraph{Convexity}
The proof for the convexity of $\mathbb{S}^{n}$ is quite simple ~\citep{dahl2010introduction}.

Suppose, $x,y \in \mathbb{S}^{n} $ and $\alpha \in [0,1]$.

Using triangle inequality~\citep{boyd2004convex},
$$\mid\mid \alpha x + (1- \alpha) y \mid\mid   \:\: \leq \:\:\mid\mid \alpha x\mid\mid + \mid\mid (1 - \alpha )y\mid\mid = \alpha + (1 - \alpha) = 1$$

Hence, $\alpha x + (1- \alpha) y \in \mathbb{S}^{n}$ and thus $\mathbb{S}^{n}$ is convex.

\paragraph{Connectedness}
To prove this, $\mathbb{S}^{n}$ is equated as union of closed upper hemisphere ($\mathbb{S}_{+}^n$) and lower hemisphere ($\mathbb{S}_{-}^n$). Individual hemispheres, $\mathbb{S}_{+}^n$ and $\mathbb{S}_{-}^n$ are homeomorphic to closed unit disk ($\mathbb{D}^{n}$) in $\mathbb{R}^n$ and intersect mutually.

First, let us consider, closed upper hemisphere $\mathbb{S}_{+}^n := \{ u \in \mathbb{R}^{n+1} : u_{n+1} \geq 0 \} $  and the unit disk $\mathbb{D}^{n} := \{ v \in \mathbb{R}^{n} :\:\: \mid\mid v \mid\mid \leq 1 \}$. We claim that, $\mathbb{D}^{n}$ shows homeomorphism with $\mathbb{S}_{+}^n$. The map $f_{+}:\mathbb{D}^{n} \to \mathbb{S}_{+}^n$ is clearly bijective and continuous. Where,

$$f_{+}(v) := \left(v_1, v_2, ...,v_n, \sqrt{1-\mid\mid v\mid\mid^2}\right) $$

As, $f_{+}$ is a continuous map from a compact topological space to a definite topological space with homeomorphism. So, in any of the case, $\mathbb{S}_{+}^n$ is connected. \\

Similarly, it is concluded that, the closed lower hemisphere $\mathbb{S}_{-}^n := \{ u \in \mathbb{R}^{n+1} : u_{n+1} \leq 0 \} $ is an image of $f_{-}:\mathbb{D}^{n} \to \mathbb{S}_{-}^n$ given by,

$$f_{-}(v) := \left(v_1, v_2, ...,v_n, -\sqrt{1-\mid\mid v\mid\mid^2}\right) $$

The observation is clear that the intersection $\mathbb{S}_{+}^n \cap \mathbb{S}_{-}^n := \{ u \in \mathbb{R}^{n+1} : u_{n+1} = 0 \}$ is a non-empty set. Hence, from theorem. 5 (Appendix.~\ref{apd:first}), we can conclude that $\mathbb{S}^{n}$ is connected.

\end{proof}

\paragraph{Proposition 1.}
\textit{A two-layered neural networks decision region possessing the property a) convex and b) connected aids in decision making by assigning decision boundaries} ~\citep{wieland1987geometric}.\textit{ As unit hypersphere possess these two properties, and it can be used as a decision region.}

\begin{proof}
Theorem~\ref{theorem1} gives sufficient justification for the above proposition 1.
\end{proof}

Thus, $ \varphi(x)$ transformation applied at the final layer of the neural networks shifts the decision regions in Euclidean space to the hyperspherical manifold. As a note, erstwhile research assumed that data points lie in hyperspherical manifold and devised objective functions. This work removed this assumption and transformed the data onto a hyperspherical manifold. For a better intuition, PyTorch-like pseudo-code is explained in Algorithm 1.


\begin{figure*}[!t]
\begin{center}
\includegraphics[width=.49\textwidth,scale=1.5]{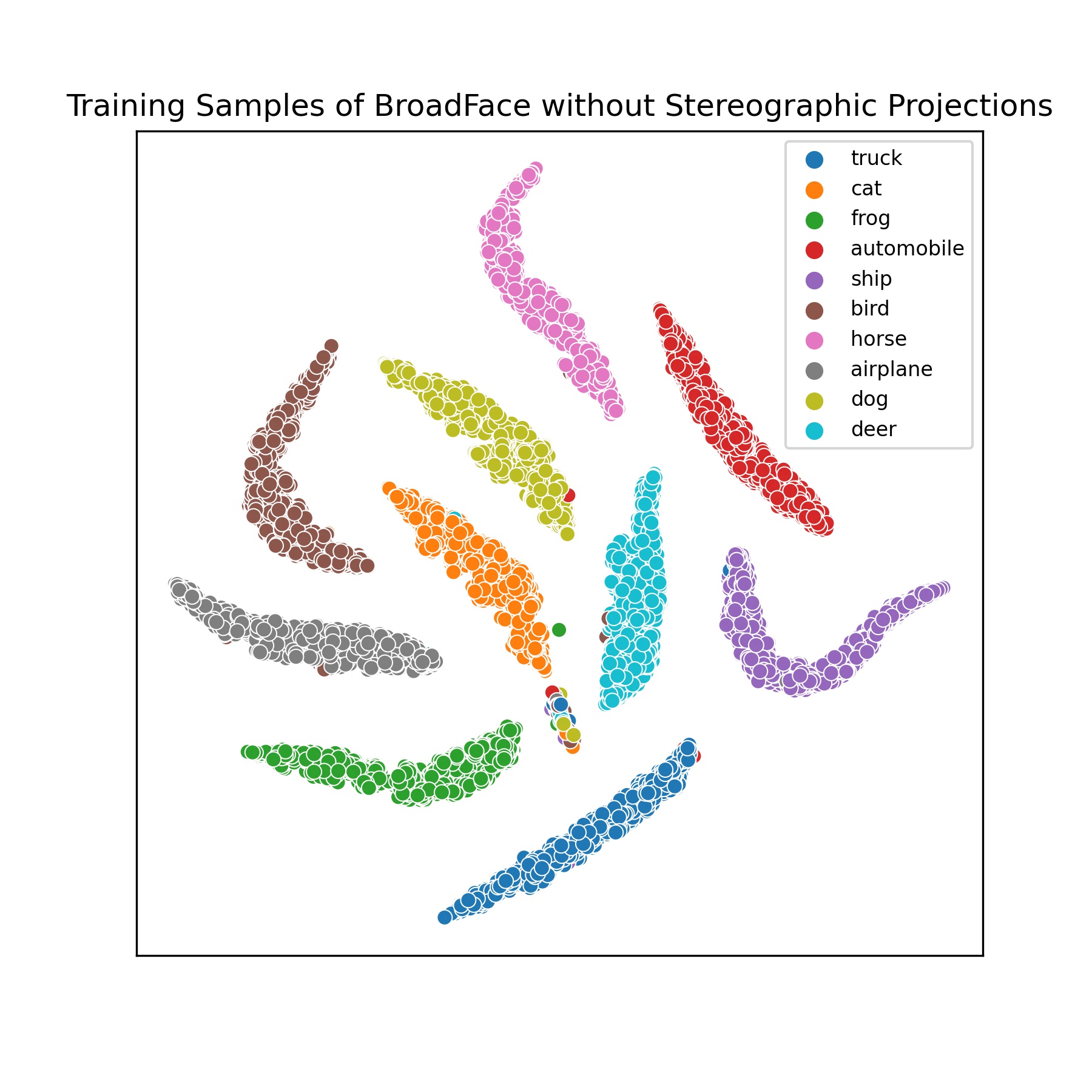}
\includegraphics[width=.49\textwidth,scale=1.5]{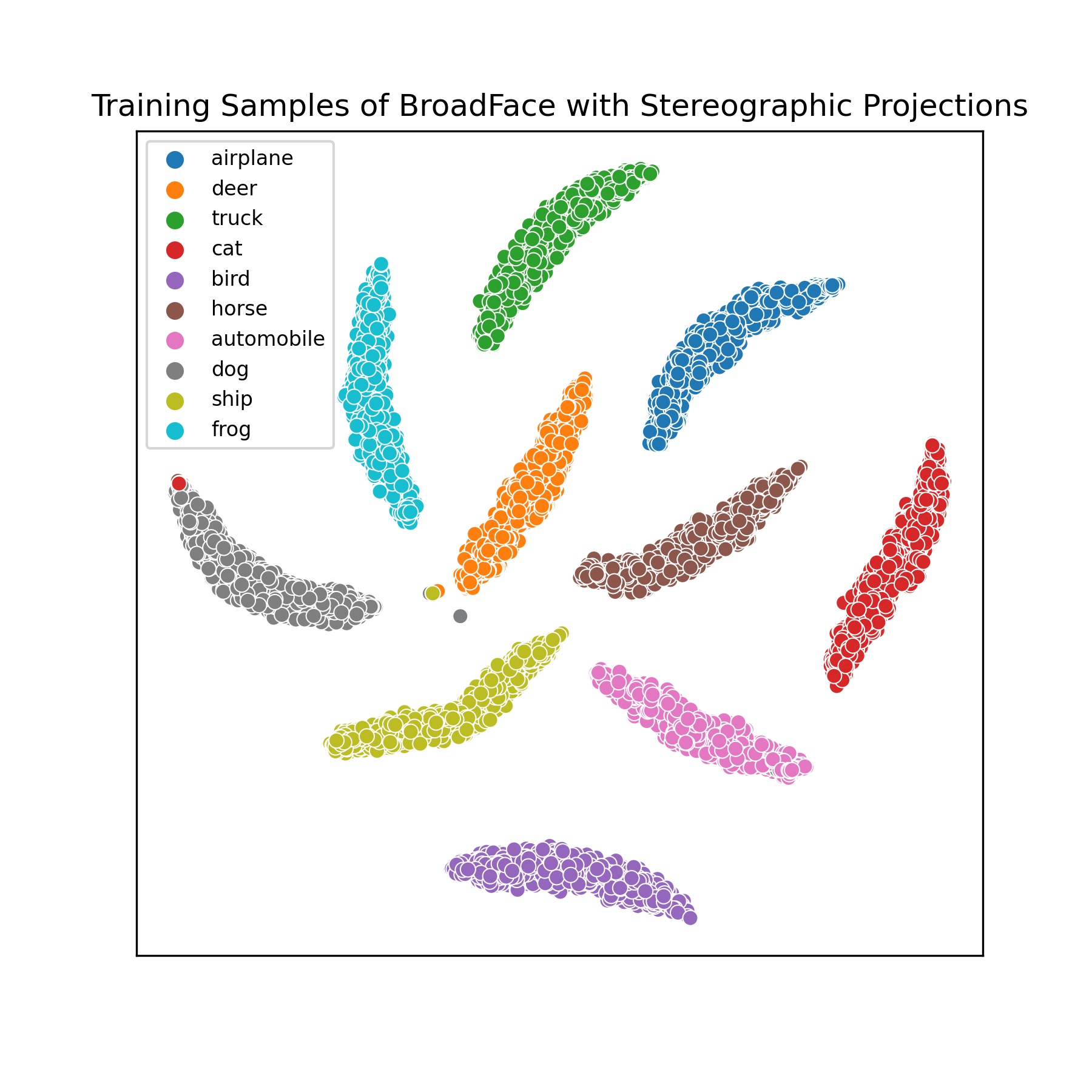}
\includegraphics[width=.49\textwidth,scale=1.5]{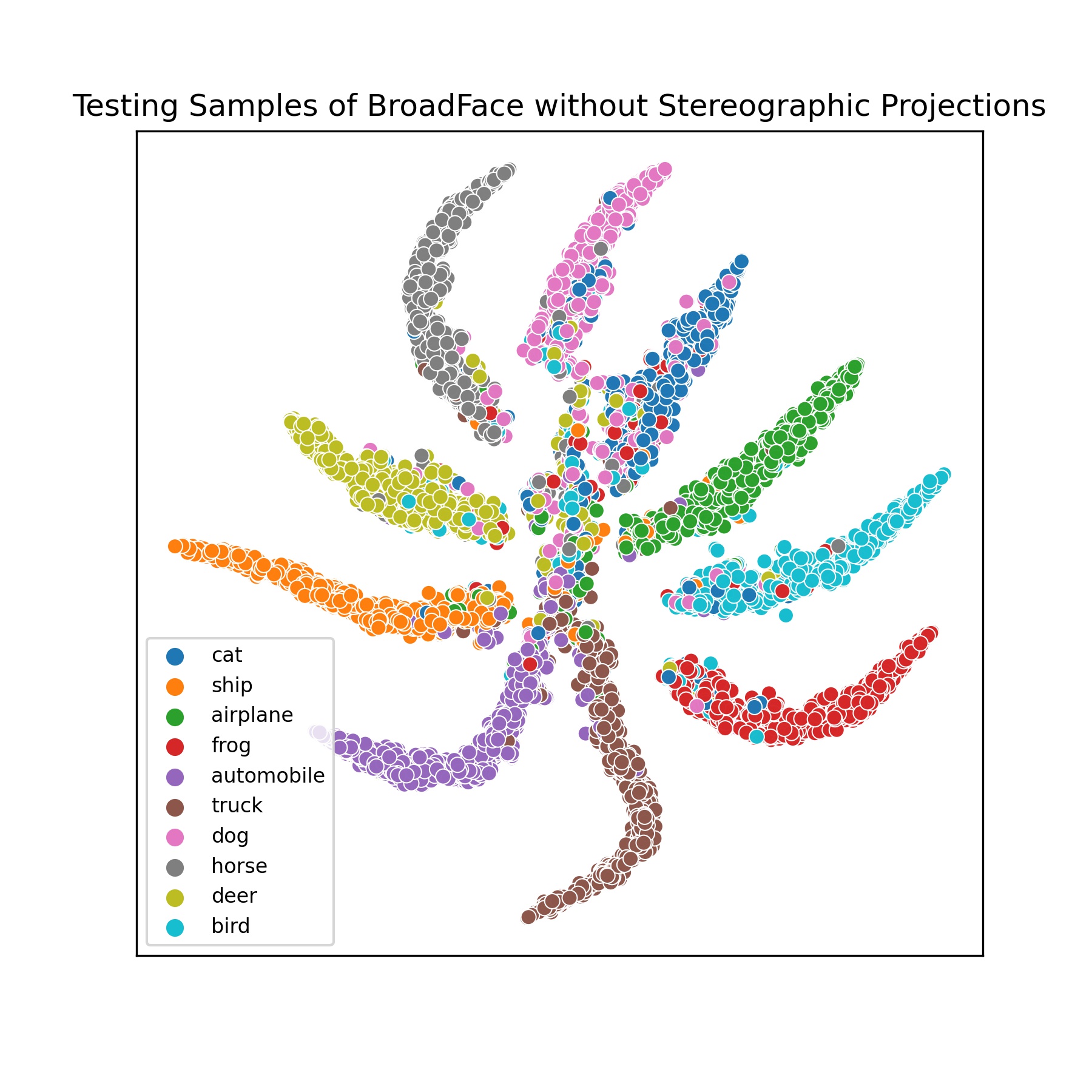}
\includegraphics[width=.49\textwidth,scale=1.5]{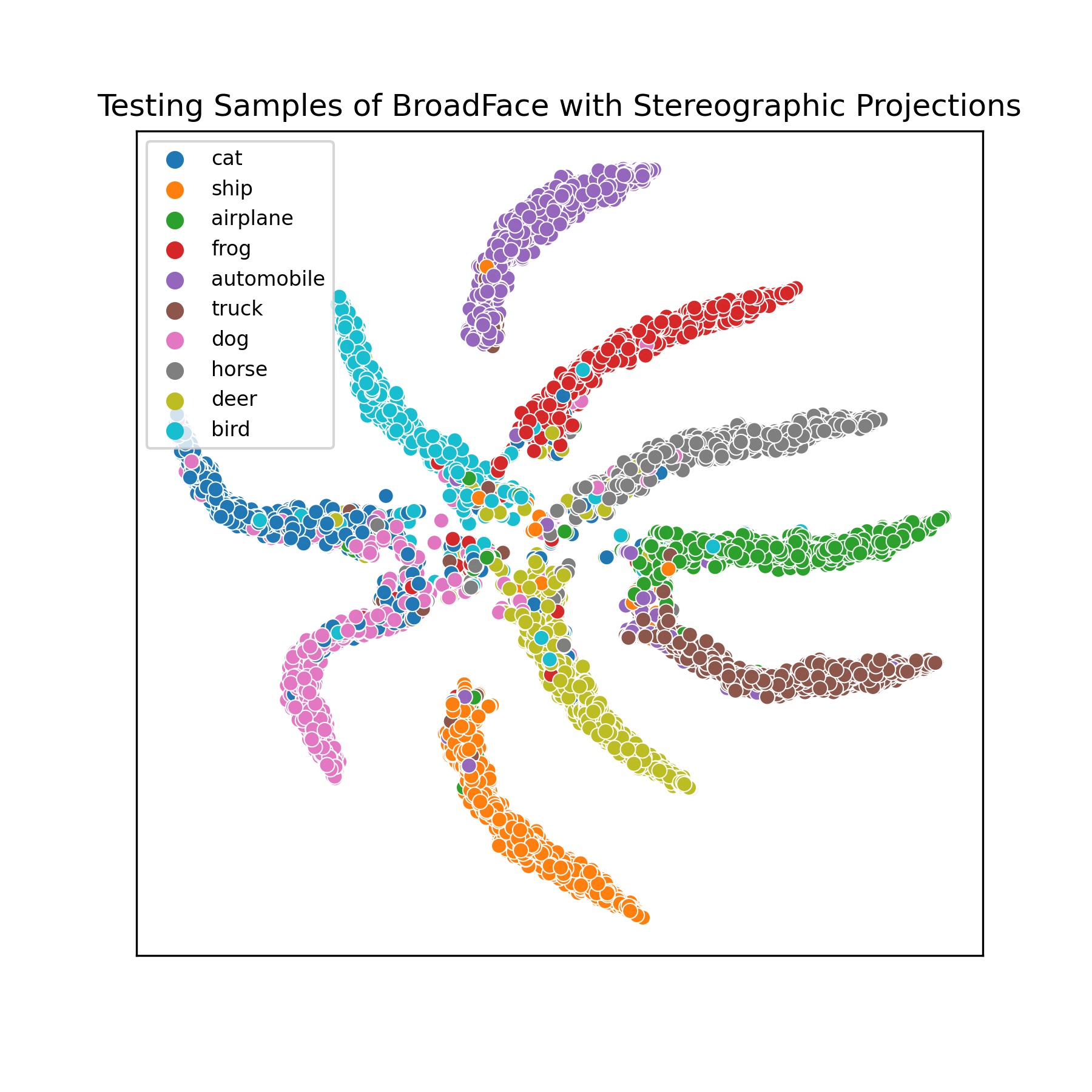}

\caption{The above figure illustrates the t-sne embedding of the final decision region (last layer of neural network). The parameters for t-sne initialized are mentioned in detail. Learning rate $\gets 200$; perplexity $\gets 50$; iterations $\gets 1000$; angle $\gets 6\times 10^{-1}$; initialized with PCA; metric as euclidean; method for computation is chosen as barrens hut. The number of samples considered to visualize both train and test is $10^4$. In the above illustration, considering The train case without projection confirms a tiny cluster including every (nearly) class, and in a test scenario, they tend to group heavily. Whereas considering the case of applying stereographic projection, there are fewer samples forming clusters during training, and there is less affinity to form a cluster in the testing phase. } \label{fig:tsne}
\end{center}
\end{figure*}


\section{Experimentation}
In this section, a brief introduction to all the angular margin loss functions is provided, as these loss functions have an underlying assumption that the data points lie in the hypersphere. But, with stereographic projection, the data points shift their space. So, we analyze that whether stereographic projections in practicality drive incredible performance for angular margin losses. Subsequently, the hyperparameters constructed while training the neural network and the training paradigm are discussed in detail. 

\subsection{Angular Margin Objectives }
First, SphereFace has presented angular-margin Softmax loss(A-Softmax loss), which defines a
decision boundary with a controlled margin. The main task of this loss function is to increase the
posterior probability for the true label by implying the multiplicative angular-margin ${m}$ between the
learned feature ${x_i}$ and the weight vector ${W_i}$ of the true label ${y_i}$. As the margin increases
for a more precise boundary, the constrained region becomes smaller, and the learning task becomes more
problematic. In the below equation weight vector is normalized using ${L_2}$ normalization. For general
understanding, we replace ${cosine(.)}$ with ${\omega(.)}$ for feasibility and the original equation is presented in ~\citep{liu2017sphereface}.

\begin{equation}
L_{SphereFace} = \frac{1}{N} \sum_{i^N} - \log \left(\frac{e^{||x_i|| (\omega ( {m\theta_{y_i,i})})}}{e^{||x_i||(\omega(m \theta_{y_i,i}))} +\sum_{j \neq {y_i}}e^{||x_i||(\omega (\theta_{j,i}))}} \right)
\end{equation}

CosFace~\citep{wang2018cosface} has introduced additive cosine margin as a decision margin. They have normalized the weight vector ${W_i}$ and the feature vector ${x_i}$ with ${L_2}$ normalization and additionally added a
scaling factor ${s}$ which is the norm of the vector ${x}$ that is, ${s}={||x||}$. Instead of tuning the
angle space, it aims to build a smooth decision boundary and gives reinforcement to the discriminative
learning by tweaking the cosine space with the help of extra margin ${m}$. For general understanding,

\begin{table}[!t]\label{table1}
\caption{The performance analysis of stereographic projection with ResNet as encoder for CIFAR-10 and 100 Data sets. The experiments were implemented five times for variant seeds (fixed batch size). Where the values are in the mean value, and the deviation is obtained. Data Augmentation is not implied while training any of the models.}
\begin{center}
\begin{tabular}{ c|c|cc } 
\hline
\multirow{3}{4em}{Datasets} & \multirow{3}{4em}{Loss Methods} & \multicolumn{2}{c} {Accuracy(\%)}  \\
  & &  \multicolumn{2}{c} {Projection applied} \\
  & & Yes & NO \\ \hline
  \multirow{5}{4em}{CIFAR-100} & CCE &55.46$\pm$0.23	&\textbf{55.65$\pm$0.20}  \\ 
  &SphereFace&	\textbf{59.68$\pm$0.15} &	57.25$\pm$0.34  \\
  &CosFace	&	\textbf{60.71$\pm$0.40} &	57.84$\pm$0.41\\
  &ArcFace	&	\bf 59.65$\pm$0.57	   &58.48$\pm$0.42 \\
  &BroadFace	&	\textbf{61.23$\pm$0.28}	&60.29$\pm$0.47\\ \hline
  \multirow{5}{4em}{CIFAR-10} &CCE		&80.92$\pm$0.41	&\textbf{81.34$\pm$0.15}\\
&SphereFace	&	\textbf{83.05$\pm$0.33}	&80.78$\pm$0.43\\
&CosFace	&	\textbf{84.31$\pm$0.32}	&81.72$\pm$0.41\\
&ArcFace		&\textbf{84.47$\pm$0.34	}&81.83 $\pm$0.52\\
&BroadFace	&\textbf{	85.59$\pm$0.24} &	81.95$\pm$0.53\\
  \hline
\end{tabular}
\end{center}
\end{table}

\begin{equation}
L_{CosFace} = \frac{1}{N}\ \sum_i^N\ - \log \left( \frac{e^{s(\omega{(\theta_{y_i,i})-
m)}}}{{e^{s(\omega{(\theta_{y_i,i})-m})} + \sum_{j\neq{y_i}} e^{s({\omega{(\theta_{j,i}))}}}}}\right)
\label{ECF}
\end{equation}

ArcFace ~\citep{deng2019arcface} has introduced additive angular-margin, which has prominent improvement over multiplicative
angular-margin and additive cosine-margin to obtain a precise decision boundary. This margin is equal to
the geodesic distance of the normalized hypersphere ~\citep{deng2019arcface} also realized the importance of normalization of
weight vector ${W_i}$ and the feature vector ${x_i}$ and added a scaling factor similar to ~\citep{wang2018cosface}. Below is
the equation of arcface and for better understanding, we replace ${\cos(.)}$ with ${\omega(.)}$ in the
original equation presented in the equations.
\begin{equation}L_{ArcFace} = \frac{1}{N}\ \sum_i^N\ - \log\left(
\frac{e^{s(\omega{(\theta_{y_i,i}+m)})}}{{e^{s(\omega{(\theta_{y_i,i}+m}))} +
\sum_{{j=1,j\neq{y_i}}}^{n} e^{s({\omega{(\theta_{j,i}))}}}}}\right)
\label{EAF}
\end{equation}

BroadFace ~\citep{kim2020broadface} has adopted the loss function presented in ~\citep{deng2019arcface} but with an intuitive training procedure, it has achieved a notable improvement in the face recognition task. For a given ${[b_i]}$ it supposedly stores the feature embedding vectors in a queue along with their identity representative vectors ${{W_y}_i}$. Utilizing this past feature embedding vectors present in the queue, it computes loss for the ${{W_y}_i}$. So the neural network has combined loss from the classifier and the embeddings. Embedding loss is compensated to mitigate the risk from past embedding vectors arisen from ${b^-}$. To reduce the additional error, a compensation function is implemented. Below is the equation for BroadFace, where ${X}$ is the sample, ${E}$ is embedding queue, ${b_i}$ is the embedding vector,  and ${b_j^*}$ compensated embedding vector.
\begin{equation}
     L_{BroadFace} = \frac{1}{X\cup E} \left( \sum_{i\in X} l(b_i) + \sum_{j\in E} l(b_j^*)\right)
     \label{BSF}
\end{equation}

\subsection{Training and Experimentation on CIFAR-10,100}
In representation learning, the classification is achieved by extracting representation from an encoder and performing non-linear mapping with feed-forward neurons. Finally, using a logit function to classify the patterns. So, we inject stereographic projection as a transformation function to project latent feature representations onto the hypersphere. This mapping is applied just before the classification. 

During the evaluation, we considered ResNet as our standard baseline encoder ~\citep{he2016deep}. Considering all the layers would be redundant. Hence architecture pruning is applied ~\citep{chen2018shallowing}, ~\citep{zhao2019variational}. The latent feature vector, $f_v$ obtained by passing through the ResNet encoder and pruning the network by skipping two pooling layers. Hence, a feature vector $f_v$ of shape $4\times4\times512 $ is obtained for standard input shape of $32\times32\times3 $ (CIFAR-10,100 data). These feature sets were fed into a two-layered, fully connected dense network with 512, 256 neurons, respectively. The intermediate activations are typically activated with ReLU. Where $\varphi$ is stereographic projection and utilizing it with angular margin loss functions at the final layer acts as transformation and eventually acts as a decision boundary. To understand the representations, t-sne ~\citep{tsne} visualizations are provided for the best performing model (BroadFace) for the CIFAR-10 model with and without projection in Figure~\ref{fig:tsne}. The results obtained in Table 1 depict the performance of stereographic projection on angular margin. The results are obtained without any augmentation as our aim was to understand the performance with standard data. Further, the experimentation is conducted five times with random seeds, and the mean and standard deviations are reported\footnote{It should be noted that training and testing samples for CIFAR-10 and 100 are $5 \times 10^4$ and $10^4$ respectively. Hence, to plot t-SNE, $10^4$ random training samples and complete testing samples were considered}.

\begin{table}[t]\label{table2}
\caption{The performance analysis of stereographic projection with ResNet as encoder for malaria data set. The experiments were implemented five times. Where the values obtained are the mean and the standard deviation of five variant executions. Data Augmentation is not implied while training any of the models.}
\begin{center}
\begin{tabular}{ c|c|cc } 
\hline
\multirow{3}{4em}{Datasets} & \multirow{3}{4em}{Loss Methods} & \multicolumn{2}{c} {Accuracy(\%)}  \\
  & &  \multicolumn{2}{c} {Projection applied} \\
  & & Yes & NO \\ \hline
  \multirow{3}{4em}{Malaria Data} & CCE &\textbf{96.05$\pm$1.32}	&94.91$\pm$2.10  \\ 
  &SphereFace&	\textbf{96.43$\pm$2.71} &	95.48$\pm$1.49  \\
  &BroadFace	&	\textbf{96.64$\pm$2.48}	&95.86$\pm$2.45\\ 
  \hline
\end{tabular}
\end{center}
\end{table}

\subsection{Training and Experimentation on Malaria Dataset}
The dataset is acquired from the NIH repository\footnote{NIH repository (National Library of Medicine) for malaria thin blood smears: \href{https://lhncbc.nlm.nih.gov/LHC-publications/pubs/MalariaDatasets.html}{Link}} and consist of two classes: parasitized and uninfected. Each of the classes has 13,780 images of varying sizes. The complete samples are reshaped into the identical shape of $128 \times 128 \times 3$. Some of the samples from the dataset are visually depicted in Figure 3. For evaluation, the training and testing samples were partitioned into 70\% and 30\% of complete data, respectively.

Now, in a similar context, ResNet is chosen as encoder ~\citep{he2016deep} and pruning is performed very similarly to that of previous. Next, SGD is chosen as optimizer (with the momentum of $0.92$) with a learning rate of $10^{-4}$ for angular margin objectives. Whereas for CCE, the learning rate is set to $10^{-3}$. The batch size is the same as that of the previous, and the evaluation has proceeded until it reaches convergence. As per the results obtained from Table 1, we decided that CosFace and ArcFace are intermediate models, i.e., their performance is higher than SphereFace and less than that of BroadFace. Hence, to reduce redundancy, these executions are not considered for the malaria dataset. As the data is sufficiently large for both classes, we did not perform any augmentation. As previously mentioned, our goal is to understand the performance of the model with pure data. The results obtained in Table 2 prove the performance of stereographic projection on angular margin objectives and CCE.

\begin{figure}[b]
\begin{center}
\includegraphics[width=0.99\textwidth]{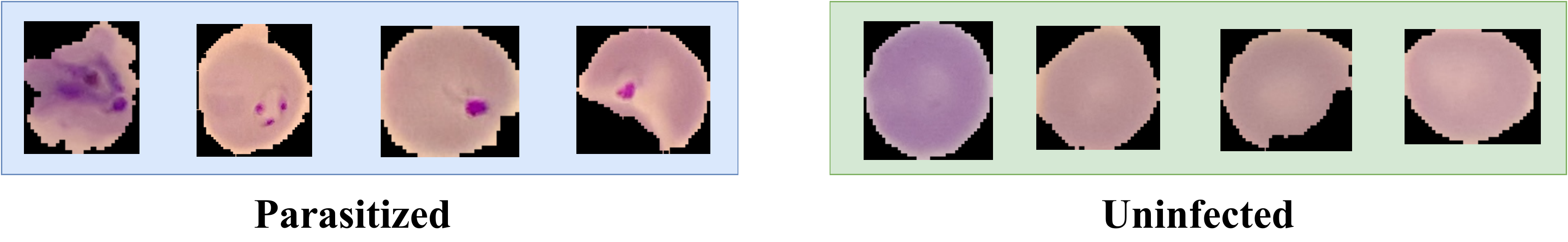}
\caption{Sample images of thin blood smear images infected with and without malaria.}
\end{center}
\end{figure}

\section{Disadvantages}
\paragraph{Additional Tuning}The hyperparameters in angular margin objective functions such as scale factor and margins are sophisticated to tune. For this, a consistent effort is required to adjust these additional hyperparameters on variant data sets.

\paragraph{Slower Training}During experimentation, we observed that execution time is slightly higher for the model with stereographic projection than that of the model without projection. This eventually consumes more electricity, and it is hazardous to the environment.

\section{Conclusion}
In this research, we justify theoretically and practically that stereographic projection integrating with angular margin objective functions for image classification problems improves the decision-making in neural networks. Instead of assuming the space, it is required to shift the space from euclidean to hypersphere to provide better performance for standard image classification data as well as malaria thin blood smear images. In the future, we aim to provide a versatile model that can enhance variant biomedical tasks' performance.

\bibliographystyle{plain}
\bibliography{references.bbl}

\begin{thebibliography}{10}

\bibitem{boyd2004convex}
Stephen Boyd, Stephen~P Boyd, and Lieven Vandenberghe.
\newblock {\em Convex optimization}.
\newblock Cambridge university press, 2004.

\bibitem{bryant1985metric}
Victor Bryant.
\newblock {\em Metric spaces: iteration and application}.
\newblock Cambridge University Press, 1985.

\bibitem{chen2018shallowing}
Shi Chen and Qi~Zhao.
\newblock Shallowing deep networks: Layer-wise pruning based on feature
  representations.
\newblock {\em IEEE transactions on pattern analysis and machine intelligence},
  41(12):3048--3056, 2018.

\bibitem{chen2020simple}
Ting Chen, Simon Kornblith, Mohammad Norouzi, and Geoffrey Hinton.
\newblock A simple framework for contrastive learning of visual
  representations.
\newblock In {\em International conference on machine learning}, pages
  1597--1607. PMLR, 2020.

\bibitem{dahl2010introduction}
Geir Dahl.
\newblock An introduction to convexity.
\newblock {\em University of Oslo, Centre of Mathematics for Applications,
  Oslo, Norway}, 2010.

\bibitem{deng2019arcface}
Jiankang Deng, Jia Guo, Niannan Xue, and Stefanos Zafeiriou.
\newblock Arcface: Additive angular margin loss for deep face recognition.
\newblock In {\em Proceedings of the IEEE/CVF Conference on Computer Vision and
  Pattern Recognition}, pages 4690--4699, 2019.

\bibitem{he2016deep}
Kaiming He, Xiangyu Zhang, Shaoqing Ren, and Jian Sun.
\newblock Deep residual learning for image recognition.
\newblock In {\em Proceedings of the IEEE conference on computer vision and
  pattern recognition}, pages 770--778, 2016.

\bibitem{NEURIPS2020_d89a66c7}
Prannay Khosla, Piotr Teterwak, Chen Wang, Aaron Sarna, Yonglong Tian, Phillip
  Isola, Aaron Maschinot, Ce~Liu, and Dilip Krishnan.
\newblock Supervised contrastive learning.
\newblock In H.~Larochelle, M.~Ranzato, R.~Hadsell, M.~F. Balcan, and H.~Lin,
  editors, {\em Advances in Neural Information Processing Systems}, volume~33,
  pages 18661--18673. Curran Associates, Inc., 2020.

\bibitem{kim2020broadface}
Yonghyun Kim, Wonpyo Park, and Jongju Shin.
\newblock Broadface: Looking at tens of thousands of people at once for face
  recognition.
\newblock In {\em European Conference on Computer Vision}, pages 536--552.
  Springer, 2020.

\bibitem{lecun2015deep}
Yann LeCun, Yoshua Bengio, and Geoffrey Hinton.
\newblock Deep learning.
\newblock {\em nature}, 521(7553):436--444, 2015.

\bibitem{liu2017sphereface}
Weiyang Liu, Yandong Wen, Zhiding Yu, Ming Li, Bhiksha Raj, and Le~Song.
\newblock Sphereface: Deep hypersphere embedding for face recognition.
\newblock In {\em Proceedings of the IEEE conference on computer vision and
  pattern recognition}, pages 212--220, 2017.

\bibitem{liu2016large}
Weiyang Liu, Yandong Wen, Zhiding Yu, and Meng Yang.
\newblock Large-margin softmax loss for convolutional neural networks.
\newblock In {\em ICML}, volume~2, page~7, 2016.

\bibitem{deephyper1}
Weiyang Liu, Yan-Ming Zhang, Xingguo Li, Zhiding Yu, Bo~Dai, Tuo Zhao, and
  Le~Song.
\newblock Deep hyperspherical learning.
\newblock In I.~Guyon, U.~V. Luxburg, S.~Bengio, H.~Wallach, R.~Fergus,
  S.~Vishwanathan, and R.~Garnett, editors, {\em Advances in Neural Information
  Processing Systems}, volume~30. Curran Associates, Inc., 2017.

\bibitem{deephyper2}
Pascal Mettes, Elise van~der Pol, and Cees Snoek.
\newblock Hyperspherical prototype networks.
\newblock {\em Advances in Neural Information Processing Systems},
  32:1487--1497, 2019.

\bibitem{park2019sphere}
Sung~Woo Park and Junseok Kwon.
\newblock Sphere generative adversarial network based on geometric moment
  matching.
\newblock In {\em Proceedings of the IEEE/CVF Conference on Computer Vision and
  Pattern Recognition}, pages 4292--4301, 2019.

\bibitem{saffery1991using}
J~Saffery and C~Thornton.
\newblock Using stereographic projection as a preprocessing technique for
  upstart.
\newblock In {\em IJCNN-91-Seattle International Joint Conference on Neural
  Networks}, volume~2, pages 441--446. IEEE, 1991.

\bibitem{tsne}
Laurens van~der Maaten and Geoffrey Hinton.
\newblock Visualizing data using t-sne.
\newblock {\em Journal of Machine Learning Research}, 9(86):2579--2605, 2008.

\bibitem{wang2018cosface}
Hao Wang, Yitong Wang, Zheng Zhou, Xing Ji, Dihong Gong, Jingchao Zhou, Zhifeng
  Li, and Wei Liu.
\newblock Cosface: Large margin cosine loss for deep face recognition.
\newblock In {\em Proceedings of the IEEE conference on computer vision and
  pattern recognition}, pages 5265--5274, 2018.

\bibitem{wang2020understanding}
Tongzhou Wang and Phillip Isola.
\newblock Understanding contrastive representation learning through alignment
  and uniformity on the hypersphere.
\newblock In {\em International Conference on Machine Learning}, pages
  9929--9939. PMLR, 2020.

\bibitem{wieland1987geometric}
Alexis Wieland and Russell Leighton.
\newblock Geometric analysis of neural network capabilities.
\newblock 1987.

\bibitem{yavartanoo2018spnet}
Mohsen Yavartanoo, Eu~Young Kim, and Kyoung~Mu Lee.
\newblock Spnet: Deep 3d object classification and retrieval using
  stereographic projection.
\newblock In {\em Asian conference on computer vision}, pages 691--706.
  Springer, 2018.

\bibitem{zeiler2014visualizing}
Matthew~D Zeiler and Rob Fergus.
\newblock Visualizing and understanding convolutional networks.
\newblock In {\em European conference on computer vision}, pages 818--833.
  Springer, 2014.

\bibitem{zhao2019variational}
Chenglong Zhao, Bingbing Ni, Jian Zhang, Qiwei Zhao, Wenjun Zhang, and Qi~Tian.
\newblock Variational convolutional neural network pruning.
\newblock In {\em Proceedings of the IEEE/CVF Conference on Computer Vision and
  Pattern Recognition}, pages 2780--2789, 2019.

\bibitem{zhou2019continuity}
Yi~Zhou, Connelly Barnes, Jingwan Lu, Jimei Yang, and Hao Li.
\newblock On the continuity of rotation representations in neural networks.
\newblock In {\em Proceedings of the IEEE/CVF Conference on Computer Vision and
  Pattern Recognition}, pages 5745--5753, 2019.

\end{thebibliography}

\appendix

\section*{Appendix}\label{apd:first}

In this section, we would like to introduce some basic theorems related to connectedness of a topological space. These theorems provides an interpretability of Theorem ~\ref{theorem1}.

\begin{customthm}{2}
A set $X$ in a topological space $\mathcal{T}$ is disconnected if $X$ can be written as intersection of two disjoint, open and non-empty sets. In other cases $X$ is connected.
\end{customthm}

\begin{proof}
The complete proof for this theorem is detailed and elucidated with examples ~\citep{bryant1985metric}.
\end{proof}


\begin{customthm}{3}
The topological space $\mathcal{T}$ is connected iff every continuous function $f:\mathcal{T} \to \{\pm 1\}$ is constant.
$$(or)$$
Consider a set $X$, subset of topological space $\mathcal{T}$, is said to be connected if and only if every continuous  function $f:{X} \to \{\pm 1\}$ is constant.

\end{customthm}

\begin{proof}
Assume, $\mathcal{T}$ is not connected.

Therefore, $\exists \:\: X,Y \subset \mathcal{T}$ are two disjoint and proper non-empty subsets in $\mathcal{T}$. These subsets in are both open and close in $\mathcal{T}$ and $\mathcal{T} = X \cup Y$.

So, $f:\mathcal{T} \to \{\pm1\}$ can be written as,
\begin{align*}
f(u)=\left\{\begin{array}{lll}
                +1 & ,if & u\in X \\
                -1 & ,if & u\in Y
            \end{array}\right.
\end{align*}

Then, $f:\mathcal{T} \to \{\pm 1\}$ is a continuous non-constant function. Hence, conversely, the theorem is proved.
\end{proof}


\begin{customthm}{4}
The topological space $\mathcal{T}$ has two connected subsets $ X$, $Y$ such that, $X \cap Y \neq \phi$. Then, $X \cup Y$ is connected,
\end{customthm}

\begin{proof}
Consider, $z \in X\cap Y$ and $f:X\cap Y \to \{\pm1\}$ be a continuous function.

$\because X$ is connected, $f$ is constant on $X$. 

So, $f(x) = f(z) \: \forall \: x \in X$.\\

Similarly, $f$ is constant on $Y$.

So, $f(y) = f(z) \: \forall \: y \in Y$.\\

From theorem. 4, it can be said that, any continuous function from $X\cup Y$ to $\{\pm1\}$ is constant. Hence, $X \cup Y$ is connected.
\end{proof}






\end{document}